\documentclass{elsarticle}

\usepackage{covington}
\usepackage{rotating}
\usepackage{graphicx}
\usepackage{multirow}

\begin{document}

\title{Analysis of Named Entity Recognition\\ and Linking for Tweets}

\author[usfd]{Leon Derczynski}
\ead{leon@dcs.shef.ac.uk}
\author[usfd]{Diana Maynard}
\author[eur,to]{Giuseppe Rizzo}
\author[vu]{Marieke van Erp}
\author[usfd]{Genevieve Gorrell}
\author[eur]{Rapha\"{e}l Troncy}
\author[usfd]{Johann Petrak}
\author[usfd]{Kalina Bontcheva}

\fntext[fn1]{To appear in ``Information Processing and Management".}

\address[usfd]{University of Sheffield, Sheffield, S1 4DP, UK}
\address[eur]{EURECOM, 06904 Sophia Antipolis, France}
\address[vu]{VU University Amsterdam, 1081 HV Amsterdam, The Netherlands}
\address[to]{Universit\`{a} di Torino, 10124 Turin, Italy}


\begin{abstract}
Applying natural language processing for mining and intelligent information access to tweets (a form of microblog) is a challenging, emerging research area. 
Unlike carefully authored news text and other longer content, tweets pose a number of new challenges, due to their short, noisy, context-dependent, and dynamic nature. 
Information extraction from tweets is typically performed in a pipeline, comprising consecutive stages of language identification, tokenisation, part-of-speech tagging, named entity recognition and entity disambiguation (e.g. with respect to DBpedia). 
In this work, we describe a new Twitter entity disambiguation dataset, and conduct an empirical analysis of named entity recognition and disambiguation, investigating how robust a number of state-of-the-art systems are on such noisy texts, what the main sources of error are, and which problems should be further investigated to improve the state of the art.
\end{abstract}


\maketitle

\interfootnotelinepenalty=8000

\section{Introduction}

Information Extraction (IE) \cite{Car97,App99} is a form of natural language analysis, which takes textual content as input and extracts fixed-type, unambiguous snippets as output. 
The extracted data may be used directly for display to users (e.g. a list of named entities mentioned in a document), for storing in a database for later analysis, or for improving information search and other information access tasks.

Named Entity Recognition (NER) is one of the key information extraction tasks, which is concerned with identifying names of entities such as people, locations, organisations and products. 
It is typically broken down into two main phases: \emph{entity detection} and \emph{entity typing} (also called classification)~\cite{grishman:1996}.
A follow-up step to NER is Named Entity Linking (NEL), which links entity mentions within the same document (also known as entity disambiguation)~\cite{hirschmann:1997}, or in other resources (also known as entity resolution)~\cite{EntityLinkingRao-et-al}. 
Typically, state-of-the-art NER and NEL systems are developed and evaluated on news articles and other carefully written, longer content~\cite{Ratinov2009,EntityLinkingRao-et-al}.

In recent years, social media -- and microblogging in particular -- have established themselves as high-value, high-volume content, which organisations increasingly wish to analyse automatically. 
Currently, the leading microblogging platform is Twitter, which has around 288 million active users, posting over 500 million tweets a day,\footnote{\texttt{http://news.cnet.com/8301-1023\_3-57541566-93 /report-twitter-hits-half-a-billion-tweets-a-day}} and has the fastest growing network in terms of active usage.\footnote{\texttt{http://globalwebindex.net/ thinking/ social-platforms-gwi-8-update-decline-of-local-social-media-platforms}}
\sloppypar{}
Reliable entity recognition and linking of user-generated content is an enabler for other information extraction tasks (e.g. relation extraction), as well as opinion mining~\cite{maynard2012challenges}, and summarisation~\cite{rout2013summarisation}. It is relevant in many application contexts~\cite{derczynski2013a}, including knowledge management, competitor intelligence, customer relation management, eBusiness, eScience, eHealth, and eGovernment.

Information extraction over microblogs has only recently become an active research topic~\cite{rowe2013msm}, following early experiments which showed this genre to be extremely challenging for state-of-the-art algorithms~\cite{derczynski2013microblog}. 
For instance, named entity recognition methods typically have 85-90\% accuracy on longer texts, but 30-50\% on tweets~\cite{ritter2011named,liujoint}. 
First, the shortness of microblogs (maximum 140 characters for tweets) makes them hard to interpret. 
Consequently, ambiguity is a major problem since semantic annotation methods cannot easily make use of coreference information. 
Unlike longer news articles, there is a low amount of discourse information per microblog document, and threaded structure is fragmented across multiple documents, flowing in multiple directions.
Second, microtexts exhibit much more language variation, tend to be less grammatical than longer posts, contain unorthodox capitalisation, and make frequent use of emoticons, abbreviations and hashtags, which can form an important part of the meaning.
To combat these problems, research has focused on microblog-specific information extraction algorithms (e.g. named entity recognition for Twitter using CRFs~\cite{ritter2011named} or hybrid methods~\cite{vanerp2013nerdml}). 
Particular attention is given to microtext normalisation~\cite{han2011lexical}, as a way of removing some of the linguistic noise prior to part-of-speech tagging and entity recognition.


In light of the above, this paper aims to answer the following research questions:

\begin{description}
  \item[RQ1] How robust are state-of-the-art named entity recognition and linking methods on short and noisy microblog texts?
  \item[RQ2] What problem areas are there in recognising named entities in microblog posts, and what are the major causes of false negatives and false positives? 
  \item[RQ3] Which problems need to be solved in order to further the state-of-the-art in NER and NEL on this difficult text genre?
\end{description}

Our key contributions in this paper are as follows.
We report on the construction of a new Twitter NEL dataset that remedies some inconsistencies in prior data.
As well as evaluating and analysing modern general-purpose systems, we describe and evaluate two domain specific state-of-the-art NER and NEL systems against data from this genre (NERD-ML and YODIE).
Also, we conduct an empirical analysis of named entity recognition and linking over this genre and present the results, to aid principled future investigations in this important area.

The paper is structured as follows.\footnote{Some parts of Section~\ref{ssec:wholetweetlink}, Section~\ref{sec:langid}, Section~\ref{sec:pos} and Section~\ref{sec:normalisation} appeared in an earlier form in~\cite{derczynski2013microblog}.}

Section~\ref{sec:ner} evaluates the performance of state-of-the-art named entity recognition algorithms, comparing versions customised to the microblog genre to conventional, news-trained systems, and provides error analysis.

Section~\ref{sec:linking} introduces and evaluates named entity linking, comparing conventional and recent systems and techniques. Sections~\ref{sec:ner} and~\ref{sec:linking} answer \textbf{RQ1}. 

Section~\ref{sec:discussion} examines the performance and errors of recognition and linking systems, making overall observations about the nature of the task in this genre. This section addresses \textbf{RQ2}.

Section~\ref{sec:preproc} investigates factors external to NER that may affect entity recognition performance. It introduces the microblog normalisation task, compares different methods, and measures the impact normalisation has on the accuracy of information extraction from tweets, and also examines the impact that various NLP pipeline configurations have on entity recognition. 

In Section~\ref{sec:conclusion}, we discuss the limitations of current approaches, and provide directions for future work. This section forms the answer to \textbf{RQ3}.  

In this paper, we focus only on microblog posts in English, since few linguistic tools have currently been developed for tweets in other languages. 


\section{Named Entity Recognition}
\label{sec:ner}

Named entity recognition (\textbf{NER}) is a critical IE task, as it identifies which snippets in a text are mentions of entities in the real world. 
It is a pre-requisite for many other IE tasks, including NEL, coreference resolution, and relation extraction.  
NER is difficult on user-generated content in general, and in the microblog genre specifically, because of the reduced amount of contextual information in short messages and a lack of curation of content by third parties (e.g. that done by editors for newswire). 
In this section, we examine some state-of-the-art NER methods, compare their performance on microblog data, and analyse the task of entity recognition in this genre.

\subsection{Existing NER systems}\label{sec:ner-systems}

A plethora of named entity recognition techniques and systems is available for general full text (cf.~\cite{nadeau2007survey,roberts2008combining,marrero2009evaluation}).
For Twitter, some approaches have been proposed but they are mostly still in development, and often not freely available. 
In the remainder of this section, we evaluate and compare a mixture of Twitter-specific and general purpose NER tools.
We want to eliminate the possibility that poor NER on tweets is systematic -- that is, related to some particular approach, tool, or technology.
To this end, we evaluate a wide range of tools and describe their operation.
It is also important to establish factors that contribute to making an entity difficult to recognise, and so we use the results of this multilateral comparison in a later analysis

For our analyses of generic NER systems, we chose those that take different approaches and are immediately available as open source. 
The first system we evaluate is ANNIE from GATE version 8~\cite{cunningham2002}, which uses gazetteer-based lookups and finite state machines to identify and type named entities in newswire text. 
The second system is the Stanford NER system~\cite{finkel2005incorporating}, which uses a machine learning-based method to detect named entities, and is distributed with CRF models for English newswire text.

Of the NER systems available for Twitter, we chose Ritter et al.~\cite{ritter2011named}, who take a pipeline approach performing first tokenisation and POS tagging before using topic models to find named entities,
reaching 83.6\% F1 measure. 
 
In addition to these, we also include a number of commercial and research annotation services, available via Web APIs and a hybrid approach \cite{vanerp2013nerdml}, named NERD-ML, tailored for entity recognition of Twitter streams, which unifies the benefits of a crowd entity recognizer through Web entity extractors combined with the linguistic strengths of a machine learning classifier.

The commercial and research tools which we evaluate via their Web APIs are Lupedia,\footnote{\texttt{http://lupedia.ontotext.com}} DBpedia Spotlight,\footnote{\texttt{http://dbpedia.org/spotlight}} TextRazor,\footnote{\texttt{http://www.textrazor.com}} and Zemanta.\footnote{\texttt{http://www.zemanta.com}} DBpedia Spotlight and Zemanta allow users to customize the annotation task, hence we applied the following settings: \textit{i)} DBpedia Spotlight=\textit{\{confidence=0, support=0, spotter=CoOccurrenceBasedSelector, version=0.6\}}; \textit{ii)} Zemanta =\textit{\{markup limit:10\}}.
\footnote{We wanted to include AlchemyAPI, but its terms and conditions prohibit evaluation without permission, and requests for permission were not answered.}
Their evaluation was performed using the NERD framework~\cite{rizzo2012nerd} and the annotation results were harmonized using the NERD ontology.\footnote{\texttt{http://nerd.eurecom.fr/ontology/nerd-v0.5.n3}}
The NERD core ontology provides an easy alignment with the four classes used for this task.
The high-performance system reported by Microsoft Research~\cite{liujoint} is not available for evaluation, so we could not reproduce these figures.


In Table~\ref{tab:nerfeatures}, we present the main characteristics of the different NER systems, as well as the NEL systems that will be evaluated in Section~\ref{sec:linking}. 

\begin{table}
\begin{small}
\rotatebox{90}{
\begin{minipage}{\textheight}
\centering\begin{tabular}{p{1.35cm}|| p{2.7cm}| p{2.7cm}| p{2.7cm}| p{2.7cm}| p{2.7cm} }
\hline
 Feature  & ANNIE  & Stanford NER & Ritter et al. & Alchemy API & Lupedia  \\
\hline 
Approach & Gazetteers and Finite State Machines & CRF& CRF &  Machine Learning & Gazetteers and rules \\
 Languages  & EN, FR, DE, RU, CN, RO, HI  & EN & EN  & EN, FR, DE, IT, PT, RU, ES, SV  & EN, FR, IT \\
 Domain  & newswire & newswire & Twitter &  Unspecified & Unspecified  \\
 \# Classes & 7 & 4, 3 or 7 & 3 or 10 & 324 &319 \\
Taxonomy & (adapted) MUC & CoNLL, ACE & CoNLL, ACE & Alchemy & DBpedia \\
Type &  Java (GATE module) & Java  & Python &Web Service &Web Service \\
License & GPLv3 & GPLv2 & GPLv3 &Non-Commercial & Unknown   \\
Adaptable & yes & yes & partially & no & no   \\
\hline
\hline
& DBpedia Spotlight & TextRazor & Zemanta & YODIE & NERD-ML  \\
\hline
Approach& Gazetteers and Similarity Metrics & Machine Learning & Machine Learning  &  Similarity Metrics & SMO and K-NN and Naive Bayes  \\
Languages& EN & EN, NL, FR, DE, IT, PL, PT, RU, ES, SV  & EN & EN   &  EN   \\
Domain& Unspecified &Unspecified  & Unspecified &  Twitter & Twitter  \\
\# Classes& 320 & 1,779   & 81  &  1,779 &  4   \\
Taxonomy & DBpedia, Freebase, Schema.org & DBpedia, Freebase & Freebase & DBpedia & NERD\\
Type& Web Service & Web Service  & Web Service& Java (GATE Module)  & Java, Python, Perl, bash \\
License & Apache License 2.0 & Non-Commercial   & Non-Commercial  &  &   GPLv3  \\
Adaptable & yes & no  & no  & yes  &  partially    \\
\end{tabular}
\caption{Key features of the different NER and NEL systems that are compared in this paper. For each of the system we indicate what type of approach is used, what languages are supported, which domain (if known), the number of classes the entity types are categorised into, how the system can be used (downloadable or for example through a Web service), which license applies to the system and whether the system can be adapted. Note that for AlchemyAPI, Lupedia, Saplo, Textrazor and Zemanta it is not public what algorithms and resources are used exactly.}
\label{tab:nerfeatures}
\end{minipage}
}
\end{small}
\end{table}

\subsection{Comparative Evaluation of Twitter NER}

\begin{table}
\centering
\footnotesize
\begin{tabular}{cc}
\hline
\textbf{Ritter} & \textbf{Stanford and ANNIE} \\
\hline
company		& Organisation \\
facility	& Location \\
geo-loc		& Location \\
movie		& Misc \\
musicartist	& Misc \\
other		& Misc \\
person		& Person \\
product		& Misc \\
sportsteam	& Organisation \\
tvshow		& Misc \\
\hline
\end{tabular}
\caption{Mappings from Ritter entity types to the Stanford and ANNIE named entity categories. Most ``musicartist'' annotations referred to either groups or were of indeterminate size, so Misc was chosen for this category instead of Person or Organisation.}
\label{tab:ritter-to-stanford-nes}
\end{table}

We evaluated the tools described above on three available datasets.
The first is the corpus of tweets developed by \cite{ritter2011named}. 
This corpus consists of 2,400 tweets (comprising 34K tokens) and was randomly sampled.
Tweet IDs are not included in this dataset, making it difficult to determine the nature of the sampling, including the relevant time window.
Examining datestamps on image URLs embedded in the corpus suggest that it was collected during September 2010.
The second is the gold standard data created through a Twitter annotation experiment with crowdsourcing at UMBC~\cite{Finin2010} (441 tweets comprising $7,037$ tokens).
The third is the dataset developed for the Making Sense of Microposts 2013 Concept Extraction Challenge~\cite{rowe2013msm}, consisting of a training and test set. 
For our evaluations, we use the test set ($4,264$ tweets comprising $29,089$ tokens). 
These datasets are all anachronistic to some degree, making them susceptible to entity drift, a significant problem in tweet datasets that we touch on in more detail in Section~\ref{wordlevelnel}.
Strict matching is used to determine scores.
Due to the short document lengths, single entity mistakes can lead to large changes in macro-averaged scores, and so we use micro-averaging; that is, scores are calculated where each entity has equal weight, instead of weighting at document level.

These datasets use disparate entity classification schemes, which we mapped to person, location, organisation, and miscellaneous using the mappings shown in Table~\ref{tab:ritter-to-stanford-nes}. 
It also needs to be noted that the different Twitter NER approaches that we evaluate and compare are trained and evaluated on small and custom datasets. 
This complicates carrying out a comparative evaluation and establishing the state-of-the-art in Twitter NER performance. 

We exclude Ritter's T-NER system from the Ritter corpus evaluation, as the released version of the system used this data for training and development. 

\begin{table}
\centering
\footnotesize
\begin{tabular}{lrrrrrrr}
\hline
	& \multicolumn{4}{c}{\textbf{Per-entity F1}} & \multicolumn{3}{c}{\textbf{Overall}}\\
\textbf{System} & \textbf{Location} & \textbf{Misc} & \textbf{Org} & \textbf{Person} &  \textbf{P} & \textbf{R} & \textbf{F1} \\
\hline
ANNIE				& 40.23	& 0.00 	& 16.00	& 24.81	& 36.14	& 16.29	& 22.46 \\
DBpedia Spotlight	& 46.06	& 6.99	& 19.44	& 48.55	& 34.70	& 28.35	& 31.20 \\
Lupedia				& 41.07	& 13.91	& 18.92	& 25.00	& 38.85	& 18.62	& 25.17 \\
NERD-ML				& {\bf 61.94}	& 23.73	& {\bf 32.73}	& {\bf 71.28}	& 52.31	& {\bf 50.69} & {\bf 51.49} \\
Stanford			& 60.49	& {\bf 25.24}	& 28.57	& 63.22	& {\bf 59.00}	& 32.00	& 41.00 \\
Stanford-Twitter	& 60.87	& 25.00	& 26.97	& 64.00	& 54.39	& 44.83	& 49.15 \\
TextRazor			& 36.99	& 12.50	& 19.33	& 70.07	& 36.33	& 38.84	& 37.54 \\
Zemanta				& 44.04	& 12.05	& 10.00 & 35.77	& 34.94	& 20.07	& 25.49 \\
\hline
\end{tabular}
\caption{Named entity recognition performance over the evaluation partition of the Ritter dataset.}
\label{tab:ner-ritter}
\end{table}

\begin{table}
\centering
\footnotesize
\begin{tabular}{ll|ccccc}
& & \multicolumn{5}{c}{ {\bf Gold} }			\\
&	Tokens&			Loc.	& Misc. &	Org.	& Person & O \\
\hline
& 	Location		&65	&3	&8	&5 	&28\\
& 	Misc			&9	&42	&14	&6 	&72\\
{\bf Response}& 	Organization	&5	&2	&18	&2 	&27\\
& 	Person			&9	&6	&6	&87 	&34\\
& 	O				&20	&25	&26	&14 	&8974\\
\end{tabular}
\caption{Confusion matrix for Stanford NER performance over the evaluation partition of the Ritter dataset.}
\label{tab:matrix-ritter}
\end{table}

\subsection{Results}

Results on the Ritter dataset are shown in Table~\ref{tab:ner-ritter}. 
We can see that conventional tools (i.e., those trained on newswire) perform poorly in this genre, and thus microblog domain adaptation is crucial for good NER. 
However, when compared to results typically achieved on longer news and blog texts, state-of-the-art in microblog NER is still lacking. 
Consequently, there is a significant proportion of missed entity mentions and false positives. 

A confusion matrix at token-level is also included, in Table~\ref{tab:matrix-ritter}.
This gives an indication of the kinds of errors made by a typical NER system when applied to tweets.

There are a number of reasons for the low results of the systems on the Ritter dataset. Partly, this is due to the varying annotation schemes and corpora on which the systems were trained. 
The annotation mapping is not perfect: for example, we mapped Facility to Organisation, but some systems will be designed to represent Facility as Location, in some or all cases. 
Similarly, some systems will consider MusicArtist as a kind of Person, but in our mapping they are Misc, because there are also bands. 
All this means that, as is common, such a comparison is somewhat imperfect and thus the comparative results are lower than those usually reported in the system-specific evaluations. 
It should also be noted that this is also a single-annotator corpus, which has implications for bias that make it hard to discern statistically significant differences in results~\cite{sogaard2014pvalue}.

\begin{table}
\centering
\footnotesize
\begin{tabular}{lrrrrrr}
\hline
	& \multicolumn{3}{c}{\textbf{Per-entity F1}} & \multicolumn{3}{c}{\textbf{Overall}}\\
\textbf{System} & \textbf{Location} & \textbf{Org} & \textbf{Person} &  \textbf{P} & \textbf{R} & \textbf{F1} \\
\hline
ANNIE				& 24.03		& 10.08		& 12.00	& 22.55	& 13.44	& 16.84	\\
DBpedia Spotlight	& 0.00		& 0.75		& 0.77	& 28.57	& 0.27	& 0.53	\\
Lupedia				& 28.70		& 19.35		& 14.21	& 54.10	& 12.99	& 20.95	\\
NERD-ML				& 43.57		& 21.45		& {\bf 49.05}	& 51.27	& {\bf 31.02} & 38.65	\\
Ritter T-NER		& 44.81		& 14.29		& 41.04 & 51.03 & 26.64 & 35.00 \\
Stanford			& {\bf 48.58}		& 27.40		& 43.07	& {\bf 64.22}	& 29.33	& \textbf{40.27}	\\
Stanford-Twitter	& 38.93		& {\bf 30.93}		& 29.55	& 38.46	& 26.57	& 31.43	\\
TextRazor			& 9.49		& 13.37		& 20.58	& 38.64	& 9.10	& 14.73	\\
Zemanta				& 35.87		& 14.56		& 11.05	& 63.73	& 12.80	& 21.31	\\
\hline
\end{tabular}
\caption{Named entity recognition performance over the gold part of the UMBC dataset.}
\label{tab:ner-umbc}
\end{table}

\begin{table}
\centering
\footnotesize
\begin{tabular}{ll|cccc}
& & \multicolumn{4}{c}{ {\bf Gold} }			\\
\hline&Tokens&	Location	&Organization	&Person &O\\
&Location		&68	&3	&6 	&152 \\
{\bf Response} &Organization	&8	&15	&10 	&229 \\
&Person			&9	&7	&33 	&207 \\
&O				&21	&11	&21 	&6150 \\
\end{tabular}
\caption{Confusion matrix for Stanford NER performance over the UMBC dataset.}
\label{tab:matrix-umbc}
\end{table}

\begin{table}
\centering
\footnotesize
\begin{tabular}{lrrrrrr}
\hline
	& \multicolumn{3}{c}{\textbf{Per-entity F1}} & \multicolumn{3}{c}{\textbf{Overall}}\\
\textbf{System} & \textbf{Location} & \textbf{Org} & \textbf{Person} &  \textbf{P} & \textbf{R} & \textbf{F1} \\
\hline
ANNIE				& 47.83		& 25.67		& 72.64		& 65.98		& 62.07		& 63.97 \\
DBpedia Spotlight 	& 35.23 	& 27.39   	& 69.11     & 59.27		& 50.42		& 54.49 \\
Lupedia				& 33.54 	& 21.30 	& 62.14 	& 65.26 	& 39.11 	& 48.91 \\
NERD-ML				& \bf{64.08} 	& {\bf 50.36} 	& {\bf 86.74} 	& 79.52 & {\bf 74.97} & {\bf 77.18}  \\
Ritter T-NER		& 43.66		& 13.73		& 83.78		& 76.25		& 65.75		& 70.61 \\
Stanford 			& 61.84		& 33.94		& 84.90		& {\bf 81.12}		& 67.23 	& 73.52	\\
Stanford-Twitter	& 50.28 	& 41.95 	& 80.00 	& 80.20 	& 63.35 	& 70.78 \\
TextRazor			& 26.13 	& 28.23 	& 78.70 	& 57.82 	& 66.90 	& 62.03	\\
Zemanta				& 46.59 	&  6.62 	& 45.71 	& 29.66 	& 27.28 	& 28.42 \\
\hline
\end{tabular}
\caption{Named entity recognition performance over the MSM2013 dataset.}
\label{tab:ner-msm}
\end{table}
\begin{table}
\centering
\footnotesize
\begin{tabular}{ll|ccccc}
& & \multicolumn{5}{c}{ {\bf Gold} }			\\
&Tokens&	Location	&Misc.	&Organisation&	Person&	O\\
\hline
&Location	&58	&3	&4	&15	&53 \\
&Misc.	&4	&40	&3	&6	&167 \\
{\bf Response}&Organisation	&12	&11	&105	&20	&180 \\
&Person	&4	&4	&11	&1697	&318 \\
&O		&21	&39	&40	&152	&26122 \\
\end{tabular}
\caption{Confusion matrix for Stanford NER performance over evaluation part of the MSM2013 dataset.}
\label{tab:matrix-msm}
\end{table}

Performance over the UMBC data is shown in Tables~\ref{tab:ner-umbc} and~\ref{tab:matrix-umbc}.
Scores here are generally lower than those over the Ritter data.
Some were particularly low; we attribute this to a high frequency of references to very transient entities such as pop music bands in the corpus.
As the corpus was designed for use as a gold standard for measuring worker reliability in crowdsourced annotation, some particularly difficult entities may have been deliberately preferred during dataset construction.
The custom-trained Stanford model achieved high scores for identifying Organisation entities again, though interestingly, the standard newswire-trained model had the best overall F1 score. The NERD-ML chain achieved the best recall score, though the fact that this value is only 31.02\% indicates how difficult the dataset is.

Finally, performance over the MSM2013 dataset is given in Tables~\ref{tab:ner-msm} and~\ref{tab:matrix-msm}. 
This is also the reason that ``MISC'' entities are not scored, because no deterministic mapping can be made from this entity class to others.
The NERD-ML approach had the overall best performance on all three stock entity types, and the Stanford NER system achieved best precision.

This particular annotation scenario, of examining only the text part of individual tweets, is sub-optimal.
Microblog users often rely on sources of context outside the current post, assuming that perhaps there is some shared relationship between them and their audience, or temporal context in the form of recent events and recently-mentioned entities.
Looking at the text in isolation removes this context, making the entity recognition harder than it would be to the actual intended reader of the message.

\subsection{Analysis}

The kinds of entities encountered in microblog corpora are somewhat different from those in typical text.
We subjectively examined the entities annotated as people, locations and organisations in the microblog corpora and the CoNLL NER training data~\cite{tjong2003introduction}.
For people, while those mentioned in news are often politicians, business leaders and journalists, tweets talk about sportsmen, actors, TV characters, and names of personal friends.
The only common type is celebrities.
For locations, news mentions countries, rivers, cities -- large objects -- and places with administrative function (parliaments, embassies). 
Tweets on the other hand discuss restaurants, bars, local landmarks, and sometimes cities; there are rare mentions of countries, often relating to a national sports team, or in tweets from news outlets.
Finally, for organisations, the news similarly talks about organisations that major in terms of value, power or people (public and private companies and government organisations) while tweets discuss bands, internet companies, and sports clubs.
Tweets also have a higher incidence of product mentions than the news genre, occurring in around 5\% of messages.

\begin{figure}
\centering
\includegraphics[width=\textwidth]{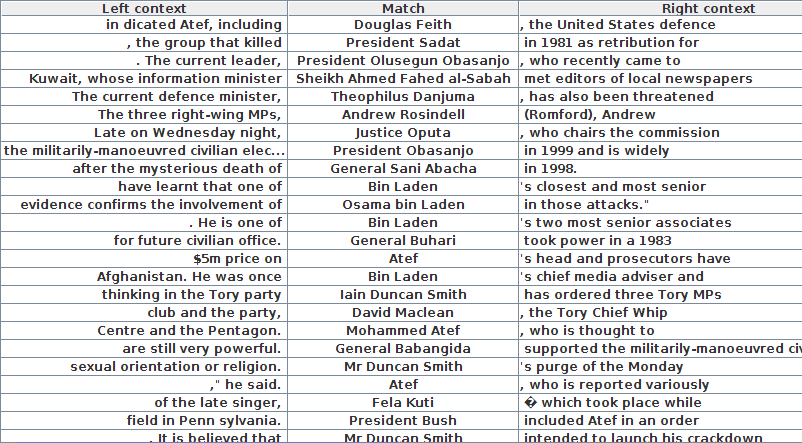}
\caption{Sentence context surrounding person entities in a sample of newswire text}
\label{fig:news-ne-context}
\end{figure}

\begin{figure}
\centering
\includegraphics[width=\textwidth]{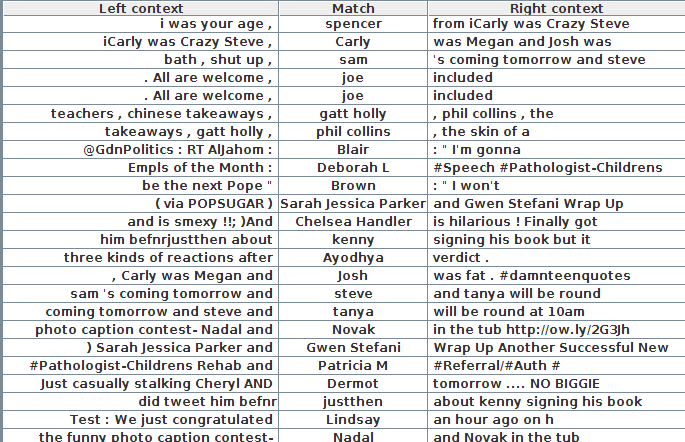}
\caption{Sentence context surrounding person entities in a sample of English tweets}
\label{fig:tweet-ne-context}
\end{figure}

That the entities occurring in tweets are different from those in newswire makes it hard for systems to tag them correctly.
Ratinov and Roth~\cite{Ratinov2009} point out that given the huge potential variety in named entity expressions, unless one has excellent generalised models of the context in which entities are mentioned, it becomes very difficult to spot previously unseen entities.
This applies to gazetteer-based approaches in particular, but also to statistical approaches.
Twitter is well-known as being a noisy genre, making it hard even for systems with perfect models of entity context to recognise unseen NEs correctly.
For example, in newswire, person entities are often mentioned by full name, preceded by a title, constitute the head of a prepositional phrase, or start a sentence, and are always correctly capitalised.
They are often followed by a possessive marker or an extended description of the entity (see Figure~\ref{fig:news-ne-context}).
This kind of linguistic context is well-formed and consistent, possibly having stability bolstered by journalistic writing guidelines.
In contrast, person entities in tweets are apparently stochastically capitalised, short, and occur in a variety of contexts~\cite{derczynski2014passive} -- including simply as exclamations (see Figure~\ref{fig:tweet-ne-context}).
This is a hostile tagging environment, where one will suffer if one expects the cues learned from heavily structured newswire to be present.

Recall, precision and F1 on all corpora were markedly lower than the typical 90\%-92\% seen with newswire text.
Incorporating twitter examples into the training data for the Stanford CRF system often led to a decrease in performance, suggesting that this approach, while effective for part-of-speech tagging (Section~\ref{sec:pos}), is not suitable for entity recognition.
The Ritter corpus was particularly tough, containing a very high level of linguistic noise and colloquialisms.
Performance on the MSM corpus was dominated by the NERD-ML system.
On examination, the entities in the MSM corpus tend to occur more frequently in well-formed tweets than in other corpora, and less frequently in particularly noisy tweets.
This may be a quirk of corpus creation; all the datasets are small.

For ease of comparison, we approached entity classification as a three- or four-way classification task.
This groups entities into an arbitrary (though conventional) scheme.
Other schemes for grouping entities into classes have been proposed; for example Ritter set up ten different categories, tuned to the types of entity found within tweets.
If one takes entity classification to the extreme, giving one class per unique entity, then the task becomes the same as NEL, which assigns a URI to each task.
NEL has the benefit of connecting an entity mention to a reference in linked data, which often includes not only existing category information but also often places an entity in a hierarchical type structure.

\section{Entity Linking}
\label{sec:linking}

Having determined which expressions in text are mentions of entities, a follow-up information extraction task is entity linking (or entity disambiguation). 
It typically requires annotating a potentially ambiguous entity mention (e.g. Paris) with a link to a canonical identifier describing a unique entity (e.g. \texttt{http://dbpedia.org/resource/Paris}). 
Approaches have used different entity databases as a disambiguation target (e.g. Wikipedia pages~\cite{cucerzan2007large,burman2011usfd,zheng2013entity}) and Linked Open Data resources (e.g. DBpedia \cite{mendes2011dbpedia}, YAGO \cite{Shen2012-Linden}, Freebase~\cite{zheng2013entity}).
Linking entity mentions to such resources is core to semantic web population and identifying differences among linking tools is crucial to achieving harmonious output~\cite{gangemi2013comparison}.
Many disambiguation targets have commonalities and links with each other, and it is often possible to map between them~\cite{rizzo2012nerd}.

Microblog named entity linking (NEL) is a relatively new, underexplored task.
Recent Twitter-focused experiments uncovered problems in using state-of-the-art entity linking in tweets~\cite{abel2011semantic,Meij2012-WSDM,greenwood2012replab}, again largely due to lack of sufficient context to aid disambiguation.
Others have analysed Twitter hashtags and annotated them with DBpedia entries to assist semantic search over microblog content~\cite{losch2011mapping}.
Approaches based on knowledge graphs have been proposed, in order to overcome the dearth of context problem, with some success~\cite{hakimov2012named}.

Research on NEL on microblog content has focused primarily on whole-tweet entity linking, also referred to as an ``aboutness'' task. 
The whole-tweet NEL task is defined as determining which topics and entities best capture the meaning of a microtext. 
However, given the shortness of microtext, correct semantic interpretation is often reliant on subtle contextual clues, and needs to be combined with human knowledge. 
For example, a tweet mentioning iPad makes Apple a relevant entity, because there is the implicit connection between the two. 
Consequently, entities relevant to a tweet may only be referred to implicitly, without a mention in the tweet's text.

From an implementational perspective, the aboutness task involves identifying relevant entities at whole-document level, skipping the common NER step of determining entity bounds. 
In cases where the task is to disambiguate only entities which are mentioned explicitly in the tweet text, not only do the entity boundaries need to be identified (e.g. start and end character offsets), but also a target unique entity identifier needs to be assigned to that entity mention. 
We refer to this NEL task as ``word level entity linking''. We start our NEL evaluation with word-level entity linking (Section~\ref{sec:word-level-nel}), followed by whole tweet linking (Section~\ref{ssec:wholetweetlink}).
\sloppypar{}
In our NEL evaluation we compared several systems. The first is YODIE~\cite{damljanovic2012named},\footnote{\texttt{https://gate.ac.uk/applications/yodie.html}} which links entities to DBpedia URIs using a combination of four metrics: string similarity, semantic similarity between nearby entities, word level contextual similarity, and URI frequency in Wikipedia articles. This system uses a Twitter tokeniser, POS tagger, and the ANNIE NER system that we described in earlier sections and is thus an end to end evaluation of microblog-adapted components (see Table~\ref{tab:nerfeatures} for details).
We also included the generic NER and NEL systems DBpedia Spotlight, Zemanta, and TextRazor presented in Section~\ref{sec:ner}.

\subsection{Word-level Entity Linking}\label{sec:word-level-nel}

Word-level linking aims to disambiguate expressions which are formed by discrete (and typically short) sequences of words.
This relies on accurate named entity recognition, so that the expressions to be disambiguated can be identified.
Expressions that are not recognised will not be disambiguated, meaning that this stage has a marked effect on overall disambiguation performance.

Research and evaluation of world-level entity linking on microblogs is currently hampered by the lack of a shared corpus of manually disambiguated entities. 
To the best of our knowledge, such resources are currently not available for others to use.\footnote{A corpus of 12\,245 tweets with semantic entity annotation was created by~\cite{liujoint}, but this will not be shared due to Microsoft policy and the system is not yet available.}

Consequently, one of the contributions of this work is in the new corpus of world-level linked entities. It is based on Ritter's named entity annotated dataset (see Section~\ref{sec:ner}), where entity mentions in tweet text are assigned an unambiguous DBpedia URI (Unique Resource Identifier).

\subsubsection{Word-level NEL Corpus}
\label{wordlevelnel}

To generate the corpus, 177 entity mentions were selected for disambiguation. Since paid-for mechanised labour raises issues with ``cheating" (intentional input of bad data) and annotation quality~\cite{Fort2011,Sabou2012a}, instead we recruited 10 volunteer NLP experts to annotate them. 

Each entity mention was labelled by a random set of three  volunteers.
The annotation task was performed using Crowdflower~\cite{biewald2012massive}.
Our interface design was to show each volunteer the text of the tweet, any URL links contained therein, and a set of candidate targets from DBpedia.
The volunteers were encouraged to click on the URL links from the tweet, to gain addition context and thus ensure that the correct DBpedia URI is chosen by them. 
Candidate entities were shown in random order, using the text from the corresponding DBpedia abstracts (where available) or the actual DBpedia URI otherwise. 
In addition, the options ``none of the above'', ``not an entity'' and ``cannot decide'' were added, to allow the volunteers to indicate that this entity mention has no corresponding DBpedia URI (none of the above), the highlighted text is not an entity, or that the tweet text (and any links, if available) did not provide sufficient information to reliably disambiguate the entity mention.

Annotations for which no clear decision was made were adjudicated by a fourth expert who had not previously seen the annotation.
As each entity annotation was disambiguated by three annotators, we determined agreement by measuring the proportion of annotations on which all three annotators made the same choice.
Out of the resulting 531 judgements, unanimous inter-annotator agreement occurred for 89\% of entities.
The resulting resource consisted of 182 microblog texts each containing entity mentions and their assigned DBpedia URIs; we make this resource available to the public with this paper.
A case study relating this particular crowdsourcing exercise is detailed in~\cite{bontcheva2014casestudy}.

Documents in the corpus were collected over a short period of time.
Sometimes this can cause problems, as due to entity ``drift"~\cite{masud2010addressing}, the selection of entities prevalent in social media (and other) discourse changes over time.
As a result, systems trained on one dataset may perform well over that dataset and other datasets gathered at the time, but not actually generalise well in the long run~\cite{fromreide2014crowdsourcing}.
In this case, the corpus is only used as evaluation data and not for training, and so this evaluation does not give an artificially drift-free scenario.

\subsubsection{Evaluation Results}

Evaluation was performed using direct strict matches only: that is, any entity extent mistakes or disambiguation errors led to an incorrect mark.
Future evaluation measures may be interested in both lenient NER evaluation (with e.g. overlaps being sufficient) as well as lenient entity disambiguation grading, through e.g. allowing entities connected through skos:exactMatch or even owl:sameAs to be credited as correct.
However, for this work, we stay with the strictest measure.

Results are given in Table~\ref{tab:linking}. 
Note that the recall here is not reliable, as we have ``downsampled'' the set of URIs returned so that all fall within DBpedia, and DBpedia does not contain references to every entity mentioned within the dataset. In addition, due to the specifics of the CrowdFlower NEL interface, annotators could not indicate missed entities for which no candidates have been assigned, even if such exist in DBpedia. 
Therefore, no result is placed in bold in this dataset.
We perform this downsampling so that we may find the simplest possible base for comparison.

\begin{table}
\centering
\footnotesize
\begin{tabular}{lccc}
\hline
\textbf{Name}            & \textbf{Precision}    & \textbf{Recall}    & \textbf{F1} \\
\hline
DBpedia Spotlight 		&  7.51	 & 27.18 & 11.77 \\
YODIE (simple)			&  36.08 & {\bf 42.79} & 39.15 \\
YODIE (news-trained)	&  {\bf 67.59} & 33.95 & {\bf 45.20} \\
Zemanta					& 34.16 & 28.92 & 31.32 \\
\hline
\end{tabular}
\caption{Tweet-level entity linking performance our entity-disambiguated dataset. YODIE results are for a whole-pipeline approach.}
\label{tab:linking}
\end{table}

\subsubsection{Error analysis}

As can be seen from these results, word-level NEL is a generally difficult task.
From observing the expert annotations, disambiguation is often hard even for humans to perform, due to lack of sufficient context from the tweet text alone. 

Amongst the systems compared here, YODIE performed best, which should be attributed at least partly to it using Twitter-adapted pre-processing components. 

One major source of mistakes is capitalisation. 
In general, microblog text tends to be poorly formed and typical entity indicators in English such as mid-sentence capitalisation are often absent or spuriously added to non-entities.

To demonstrate this point, we compare the proportional distribution of lower, upper and sentence-case words in a sentence with the tokens which are subject to false positives or false negatives, in Table~\ref{tab:fnfp-case}.
It can be seen that the majority of false positives are in sentence case, and the largest part of false negatives are in lowercase.
Conversely, almost no false positives are in lower case.
In both cases, very different proportions are mislabelled from the underlying distributions of case, indicating a non-random effect.

\begin{table}
\centering
\footnotesize
\begin{tabular}{lccc}
\hline
\textbf{Group} & \textbf{All capitals} & \textbf{All lowercase} & \textbf{Sentence case} \\
\hline
Mean proportion per tweet 		& 7.9\%		& 74.1\%	& 18.1\% \\
False positives					& 22.4\%	& 4.7\%		& 65.9\% \\
False negatives					& 9.3\%		& 39.8\%	& 37.3\% \\
\hline
\end{tabular}
\caption{Proportion of false negatives and false positives from Stanford NER over the Ritter dataset, grouped by word case.}
\label{tab:fnfp-case}
\end{table}

Tweet text is often short, too, and sometimes even human annotators did not have enough context to disambiguate the entities reliably.
For example, one candidate tweet mentioned ``\emph{at Chapel Hill}''.
While ``Chapel Hill" is almost certainly a location, it is a very common location name, and not easy to disambiguate even for a human.
Indeed, annotators disagreed enough about this tweet for it to be excluded from the final data.

Further, one may mention a friend's name, which is correctly tagged as a named entity; however, the friend is almost certainly not listed in the general knowledge resources such as DBpedia, and therefore cannot be disambiguated against this resource (i.e., we have no choice but to assign a ``nil'' URI, indicating that the result could not be disambiguated).
In such cases, a FOAF reference~\cite{brickley2010foaf} would perhaps be more suitable.

Also, many named entities are used in unconventional ways.
For example, in the tweet ``\emph{Branching out from Lincoln park after dark ... Hello ``Russian Navy", it's like the same thing but with glitter!}'', there are no location or organisation entities.
Note that \emph{Russian Navy} is collocated with \emph{glitter}, which is a little unusual (this could be determined through language modelling).
It refers in fact to a nail varnish colour in this context, and is undisambiguable using DBpedia or many other general-purpose linked entity repositories.
\emph{Lincoln Park After Dark} is another colour, and compound proper noun, despite its capitalisation -- it does not refer to the location, Lincoln Park.

The general indication is that contextual clues are critical to entity disambiguation in microblogs, and that sufficient text content is not present in the terse genre.
As usual, there is a trade-off between precision and recall.
This can be affected by varying the number of candidate URIs, where a large number makes it harder for a system to choose the correct one (i.e. reduces precision), and fewer candidate URIs risks excluding the correct one entirely (i.e. reduces recall).

\subsection{``Whole tweet'' Entity Linking}\label{ssec:wholetweetlink}
The most appropriate dataset to evaluate whole tweet entity linking is the topic-disambiguated  dataset detailed in~\cite{Meij2012-WSDM}. 
It contains both tweets and the corresponding Wikipedia articles, which are the topic of these tweets. 
Since each tweet is topic-annotated, there is no guarantee that the corresponding Wikipedia article is an entity (e.g. topics include ``website'', ``usability'', and ``audience''). 
Even when the assigned topics are entities, the tweets often contain other entities as well, which are not annotated.
These characteristics make measurements using this dataset only a rough indicator of entity linking performance.  

\subsubsection{Evaluation Results}
We evaluated the performance of DBpedia Spotlight, Zemanta and TextRazor on this corpus (see Table~\ref{tab:wsdm}). 
These are the systems that are geared to providing whole-tweet accuracy; including others would generate misleading results.
Only the top-ranked entities were considered for matching, and we discarded recognised but unlinked entities. 

\subsubsection{Error analysis}
DBpedia Spotlight performed quite poorly in terms of precision, probably due to the nature of the dataset, although it achieved the highest recall. 
In contrast, Zemanta performed poorly in terms of recall, but reasonably on precision, thus achieving the highest F1 score. 
TextRazor, although connected to a variety of linking resources, had the worst recall, but the highest precision. 

Given the limitations of this dataset, these results are somewhat inconclusive. The primary source of problems of using a corpus such as this one for evaluating NEL is that entities annotated at tweet level may better represent mentions of a topic describing the text, rather than actual mentions of an entity in the text. Unfortunately, there are no other, more suitable datasets currently available. 

\begin{table}
\centering
\footnotesize
\begin{tabular}{lccc}
\hline
\textbf{System} & \textbf{Precision} & \textbf{Recall} & \textbf{F1} \\
\hline
DBpedia Spotlight     	& 20.1\%    & {\bf 47.5\%}    	& 28.3\%    \\
TextRazor		& {\bf 64.6\%} & 26.9\%		& 38.0\%    \\
Zemanta         	& 57.7\%    & 31.8\%    	& {\bf 41.0\%}    \\
\hline
\end{tabular}
\caption{Performance of entity linking over the WSDM2012 tweet dataset; note low performance due to varying levels of specificity in tweet-level annotations.}
\label{tab:wsdm}
\end{table}

\subsection{Discussion}

To summarise, named entity linking in tweets is a challenging task, due to the limited surrounding context, as well as the implicit semantics of hashtags and user name mentions. 

Furthermore, comparing methods reliably is currently very difficult, due to the limited datasets available in this genre. 
As a first step towards alleviating this problem, we have created a small-size gold standard of tweets, which are expert annotated with DBpedia URI. 
We are currently working on expanding this gold standard through paid-for CrowdFlower experiments, where the current data will be used as gold standard units to help train and monitor the worker's performance. 
This is an essential step in the prevention of cheating, where annotators deliberately enter nonsense data in order to save time. 
We will also require multiple judgements per disambiguated entity, so that low quality annotations can be detected and removed in a reliable manner \cite{Sabou2012a,SABOU14.497.L14-1412}.   

In terms of method performance, one of the main sources of errors is unreliable microblog capitalisation. 
In news and similar genres, capitalisation is often a reliable indicator of a proper noun, suggesting a named entity. 
Unfortunately, tweets tend to be inconsistently capitalised and performance suffers as a result. 

This can in part be attributed to prior named entity recognition problems, in part -- to the lack of suitably entity annotated gold standard datasets, and in part -- to the short, noisy, and context-dependent nature of tweets. 
The rest of the paper discusses ways of addressing some of these problems.


\section{Analysis}
\label{sec:external}
\label{sec:discussion}

So far, we have demonstrated experimentally that extracting information from microblog text is a challenging task.
Problems lie both in named entity recognition and in entity linking, having negative impact on each subsequent task.
Some genre-adaptation has been attempted, often successfully, but far from the performance levels found on established text genres, such as news.
In this section, we discuss first why processing microblogs is so challenging, and then carry out error analysis. These lead to evidence-based suggestions for future research directions towards improved NER and NEL for microblog text.

\subsection{General Microblog Challenges}

Microblogs have a number of genre-specific characteristics, which make them particularly challenging for information extraction in general, and NER and NEL in particular. These are as follows:
\begin{description}
  \item [Short messages (microtexts):] Twitter and other microblog messages are very short. Some NEL methods supplement these with extra information and context coming from embedded URLs and hashtags. A recent study of 1.1 million tweets has found that 26\% of English tweets contain a URL, 16.6\% a hashtag, and 54.8\% contain a user name mention \cite{carter2013microblog}. For instance, YODIE and Mendes \emph{et al.} exploit online hashtag glossaries to augment tweets \cite{Mendes2010-WI}.  

   \item [Noisy content:] social media content often has unusual spelling (e.g. 2moro), irregular capitalisation (e.g. all capital or all lowercase letters), emoticons (e.g. :-P), and idiosyncratic abbreviations (e.g. ROFL, ZOMG). Spelling and capitalisation normalisation methods have been developed and will be discussed in Section~\ref{sec:normalisation}. 

    \item [Social context] is crucial for the correct interpretation of social media content. NER and NEL methods need to make use of social context (e.g. what are the names of the poster's social connections, where is the poster located), in order to disambiguate mentions of person names and locations reliably.  
    
    \item [User-generated:] since users produce as well as consume social media content, there is a rich source of explicit and implicit information about the user, e.g. demographics (gender, location, age), friendships. The challenge here is how to combine this information with knowledge from external resources, such as DBpedia, in order to achieve the best possible accuracy and coverage of the NER and NEL methods. 

   \item [Multilingual:] Social media content is strongly multilingual. For instance, less than 50\% of tweets are in English, with Japanese, Spanish, Portuguese, and German also featuring prominently \cite{carter2013microblog}. Unfortunately, information extraction methods have so far mostly focused on English, while low-overhead adaptation to new languages still remains an open issue. Automatic language identification \cite{carter2013microblog,lui2012langid} is an important first step, allowing applications to first separate social media in language clusters, which can then be processed using different algorithms.  

\end{description}   

Next, we turn to an analysis of errors and discuss how these could be addressed. 

\subsection{Prevalent Errors}

The main source of errors lies in the difficult cases, which are specific to this genre and are problematic for all stages of NER and NEL. 
We examined the datasets to find qualitative explanations for the data.
Then, for each of these, we quantify the impact of each explanation.

\subsubsection{Capitalisation}
The first error source is capitalisation, which causes problems for POS tagging, NER and entity linking.
In each case, where capitalisation is used in well-formed text to differentiate between common nouns and proper nouns, altering this information (e.g. through use of lower caps for convenience or all caps for emphasis) has led to incorrect decisions being made.
Correcting capitalisation is difficult, especially in the cases of polysemous nouns that have named entity senses (e.g. \emph{eat an \underline{apple}} vs. \emph{\underline{Apple Inc.}}, or \emph{the town \underline{Little Rock}} vs. \emph{throw a \underline{little rock}}).

\begin{table}
\centering
\footnotesize
\begin{tabular}{lrrrrrr}
\hline
 & \multicolumn{3}{c}{All words} & \multicolumn{3}{c}{Named entities only} \\
\textbf{Corpus} & \textbf{Lower} & \textbf{Sentence} & \textbf{Caps} & \textbf{Lower} & \textbf{Sentence} & \textbf{Caps} \\
\hline
Penn Treebank / WSJ	& 68.5\%	& 25.5\%	& 5.9\%	& 1.1\%		& 84.5\%	& 14.0\%	\\
Ritter				& 76.1\%	& 15.8\%	& 7.6\%	& 14.3\%	& 71.3\%	& 11.4\%	\\
MSM					& 65.9\%	& 25.5\%	& 8.2\%	& 6.4\%		& 86.0\%	& 6.2\%	\\
UMBC				& 73.4\%	& 20.2\%	& 5.7\%	& 17.3\%	& 71.6\%	& 8.8\%	\\
\hline
\end{tabular}
\caption{Capitalisation, in twitter and newswire corpora.}
\label{tab:cap-survey}
\end{table}

Two types of miscapitalisation can occur: words that should be capitalised being presented in lower case, and words that should be in lower case being presented in capitals or sentence case.
We examine the general capitalisation of words in newswire and twitter text, looking at both all words and just named entities.
Table~\ref{tab:cap-survey} shows that while the corpora in general exhibit similar proportions of lower case, sentence case (i.e. with only the first letter capital), and upper case words, entities in the newswire corpus have distinct capitalisation.
Notably, they have the lowest likelihood of being in lowercase, a high likelihood of being in sentence case, and the comparatively highest likelihood of being in capitals.
Named entities in social media text are more likely to be expressed using lower case.
Indeed, the data suggest that the MSM corpus is the closest to the clean PTB news text that NER systems are trained on. 
This goes partly towards explaining the high NER results reported in Table~\ref{tab:cap-survey}, whereas the Ritter and UMBC corpora are much harder for the NER systems.

To investigate the information conveyed by capitalisation, we compared entity recognition on a newswire dataset with performance on the same dataset with all tokens lowercased, using the default newswire-trained Stanford NER tool and part of the CoNLL dataset~\cite{tjong2003introduction}.
On the plain dataset, the tool achieves 89.2 precision and 88.5 recall.
On the lower case version, it reaches a lower precision of 81.7 and a low recall of 4.1, for an F1 of 7.8.
The performance drop is large.
This suggests that the capitalisation variations observed in tweets are damaging to named entity recognition performance.

\subsubsection{Typographic errors}
Secondly, typographic errors confuse many pre-linking stages, specifically tokenisation, POS tagging, and NER.
Added, skipped and swapped letters have all been found to cause problems.
Although normalisation offers some help, the state-of-the-art does not yet give a good enough balance between ignoring correct out-of-vocabulary (OOV) words and correctly repairing mistyped in-vocabulary (IV) words, demonstrated in the low impact that normalisation has on NER performance (see Section~\ref{sec:normalisation}).

\begin{table}
\centering
\footnotesize
\begin{tabular}{lrr}
\hline
\textbf{Corpus} & \textbf{IV} & \textbf{OOV} \\
\hline
Penn Treebank / WSJ	&	86.1\%	&	13.9\% \\
Ritter				&	70.4\%	&	29.6\% \\
MSM					&	66.5\%	&	33.5\% \\
UMBC				&	72.9\%	&	27.1\% \\
\hline
\end{tabular}
\caption{Proportion of in- and out-of-vocabulary words in newswire and twitter corpora.}
\label{tab:typo-survey}
\end{table}

To measure the extent of typographic errors in social media, we compare the OOV/IV distribution of tokens comprised of only letters within social media text and within a similarly-sized sample of newswire text, taken from the Penn Treebank.
The vocabulary used was the iSpell dictionary, having 144K entries.\footnote{\texttt{http://www.gnu.org/software/ispell}}
Results are given in Table~\ref{tab:typo-survey}.
The social media text has an OOV rate 2-2.5 times that of newswire text.
Note that the MSM corpus has the highest OOV rate of all microblog corpora, and yet NER results on that are the best of all three.

\begin{table}
\centering
\footnotesize
\begin{tabular}{lrr}
\hline
\textbf{Corpus} & \textbf{NE Recall} & \textbf{NE Precision} \\
\hline
MSM					& 53.7\%	& 63.5\% \\
\vspace{1mm}
UMBC				& 37.1\%	& 31.5\% \\
WSJ - unperturbed	& 89.2\%	& 88.5\% \\
WSJ	- perturbed		& 85.6\% 	& 84.3\% \\
\hline
\end{tabular}
\caption{Statistical NE recognition performance over social media, newswire and perturbed newswire text.}
\label{tab:typo-perturb}
\end{table}

To determine the impact that these errors have on statistical named entity recognition, we increase the OOV rate in gold-standard named entity annotated newswire text to that found in social media text.
This is done by randomly perturbing IV tokens in newswire text\footnote{via ROT13 encoding of a single random letter.} to make them OOV.
Our perturbed corpus ended up with an IV rate of 58.3\%.
The entity recognition rate is then compared on newswire, perturbed newswire and social media, using the default newswire-trained Stanford NER tool.
Data was taken from the CoNLL dataset and the four-class model used.
Results are given in Table~\ref{tab:typo-perturb}.
It can be seen that perturbing the newswire dataset's spellings has a negative impact on named entity recognition performance, though it does not account for all of the performance gap between social media and newswire.

\subsubsection{Shortenings}
Thirdly, microblog shortness encourages the use of a compressed form of language.
This leads to rare, uncommon or incomplete grammatical structures being used, as well as abbreviations, heavy pronoun use and uncommon abbreviations.
Typically, one would use linguistic context to unpack these information-dense, peculiarly-structured utterances; however, that is often not available.
This creates many problems, and perhaps the best way to overcome it is to create large and more-richly annotated resources. 

\begin{table}
\centering
\footnotesize
\begin{tabular}{lrrr}
\hline
\textbf{Corpus} & \textbf{Chars per word} & \textbf{Words per sent.} & \textbf{Slang shortenings} \\
\hline
Newswire (PTB WSJ)	& 5.1	& 9.1	& 0.0059\% \\
Ritter				& 4.1	& 14.8	& 0.85\% \\
MSM					& 4.3	& 14.0	& 0.43\% \\
UMBC				& 4.3	& 12.5	& 0.20\% \\
\hline
\end{tabular}
\caption{Word length, sentence length and slang incidence in twitter and newswire corpora.}
\label{tab:short-survey}
\end{table}

We compare word length (in characters) and sentence length (in words) across twitter and newswire corpora.
We also measure the occurrence of shortened words forms in newswire and tweets, using a slang list developed for twitter part-of-speech tagging~\cite{derczynski-EtAl:2013:RANLP-2013}.
Results are given in Table~\ref{tab:short-survey}.
Words are generally shorter in tweets. 
Sentences appear longer in tweets, which is counter-intuitive. 
It may be due to the inclusion of hashtags and URLs at the end of tweets, which are often introduced without ending the body sentence with e.g. a full stop.
However, although slang is an order of magnitude more frequent, it is still rare in tweets.

\subsubsection{Other problems}
As discussed in Section~\ref{sec:linking}, the lack of context is a particular problem in the entity linking task.
Even the best performing systems reach scores much lower than they would on well-formed text.
As with other linguistic disambiguation tasks, context is critical to resolving polysemous words.
Methods for increasing the amount of available context are therefore required.

Finally, there is a general need for larger linguistic resources for NEL and NER.
Annotated microblog text is particularly rare, and in this difficult genre, very much needed, in order to improve the performance of the supervised learning methods.

\section{Reducing Microblog Noise through Pre-Processing}\label{sec:preproc}

As discussed in Section~\ref{sec:discussion}, microblog noise is one of the key sources of NER and NEL errors. Moreover, since NER and NEL approaches tend to rely on shallow, upstream pre-processing components (e.g. language detection, tokenisation, part-of-speech tagging), mistakes due to noise tend to cascade down the pipeline. 

In this section, we investigate whether the impact of microblog noise on NER and NEL can be reduced by adapting the typical pre-processing algorithms to the specifics of this genre. Approaches to each pre-processing stage are discussed, contrasting a variety of methods, and comparing performance at each stage on microblog and longer, cleaner content (e.g. news\-wi\-re) text.

\begin{table*}
\centering
\footnotesize
\begin{tabular}{lrrrrrr}
\hline
\textbf{System} & \textbf{Overall} & \textbf{English} & \textbf{Dutch} & \textbf{French} & \textbf{German} & \textbf{Spanish} \\
\hline
TextCat 		& 89.5\%	& 88.4\%	& 90.2\%	& 86.2\%	& 94.6\%	& 88.0\%	\\
langid 			& 89.5\%	& 92.5\%	& 89.1\%	& 89.4\%	& 94.3\%	& 83.0\%	\\
Cybozu			& 85.3\%    & 92.0\%	& 79.9\%	& 85.8\%	& 92.0\%	& 77.4\%	\\
TextCat (twitter)	& {\bf 97.4\%}	& {\bf 99.4\%}	& {\bf 97.6\%}	& {\bf 95.2\%}	& {\bf 98.6\%}	& {\bf 96.2\%}	\\
langid (twitter)	& 87.7\%	& 88.7\%	& 88.8\%	& 88.0\%	& 92.5\%	& 81.6\%	\\
\hline
\end{tabular}
\caption{Language classification accuracy on the ILPS dataset for systems before and after adaptation to the microblog genre.}
\label{tab:langid}
\end{table*}

\subsection{Language Identification}
\label{sec:langid}

Since microblog content is strongly multilingual (see Section~\ref{sec:discussion}), \emph{language identification} needs to be performed prior to other linguistic processing, in order to ensure that the correct NER and NEL component is run.
Named entity recognition depends on correct language identification.
Not only do names have different spellings in different languages, but critically, the surrounding syntactic and lexical context is heavily language-dependent, being filled with e.g. names of jobs, or prepositions (Figure~\ref{fig:news-ne-context}).

TextCat~\cite{cavnar1994n} and the Cybozu language detection library~\cite{nakatani2010langdetect} rely on n-gram frequency models to discriminate between languages, relying on token sequences that are strong language differentiators.
Information gain-based langid.py~\cite{lui2012langid} uses n-grams to learn a multinomial event model, with a feature selection process designed to cope with variations of expression between text domains. Both TextCat and langid.py have been adapted for microblog text, using human-annotated data from Twitter. 
The former adaptation~\cite{carter2013microblog} works on a limited set of languages; the latter~\cite{preotiuc:ramss12} on 97 languages. 

We evaluated system runs on the ILPS TextCat microblog evaluation dataset.\footnote{\texttt{http://ilps.science.uva.nl/resources/twitterlid}}
Results are given in Table~\ref{tab:langid}, with the Twitter-specific versions marked ``twitter''. 
Microblog-adapted TextCat performed consistently better than the other systems.

The adapted version of TextCat has a slightly easier task than that for langid.py because it expects only five language choices, whereas the adapted langid.py is choosing labels from a set of 97 languages. 
The latter assigned a language outside the five available to 6.3\% of tweets in the evaluation set.
The adapted langid.py performed worse than the generic version.
The results are quite close for some languages, and so if an approximate 6\% improvement could be made in these cases, the Twitter-adapted version would be better. 
Although language identification is harder on tweets than on longer texts, its performance is sufficiently high to inform reliable choices in later stages.

\subsection{Part-of-speech Tagging}
\label{sec:pos}

Part-of-speech (POS) tagging is generally the next stage in a semantic annotation pipeline following tokenisaton, and is necessary for many tasks such as named entity recognition and linking.
Early high-performance tagging approaches include the Brill tagger, which uses transformation-based learning, and has the benefit of being fast~\cite{hepple2000independence}.
Later, Toutanova et al.~\cite{toutanova2003feature} introduced the Stanford tagger, trained on newswire texts, and which has sophisticated feature extraction, especially for unknown words, and a highly configurable re-trainable engine.
Models using the Penn Treebank (PTB) tagset~\cite{marcus1993building} are available for both these taggers.

However, these are not necessarily suitable for microblogs, and thus specific taggers have been developed to handle these.
We concentrate on the PTB tagset, as many existing NLP components expect this labelling schema, and changes in tagset hamper tagger comparison.
Ritter et al.~\cite{ritter2011named} trained on PTB-tagged tweets, adding extra tag labels for retweets, URLs, hashtags and user mentions.
Their work also included the distribution of a ground truth annotated tweet dataset.

We use the experimental setup detailed in Ritter's paper, apart from using a fixed train/test split in the Twitter data, where the evaluation portion had 2,242 tokens.
Results are given in Table~\ref{tab:pos}, including comparison against sections 22-24 of the Wall Street Journal part of the Penn Treebank~\cite{marcus1993building}.

The results for the Stanford and Brill taggers trained on newswire text show poor success rates;
it was not possible to re-train the Ritter tagger on newswire data only.
Results are also given for the taggers trained on Twitter data --  a re-trained, re-tuned Stanford tagger~\cite{derczynski-EtAl:2013:RANLP-2013}, Ritter's stock tagger,\footnote{\texttt{https://github.com/aritter/twitter\_nlp}} and the CMU ARK tagger v.0.3.2 with Ritter's extended PTB tagset.
Even the performance of the worst tagger trained with microblog text was better than the best tagger not using microblog data.
Ritter's system did not provide information about unknown word tagging accuracy.

\begin{table}
\centering
\footnotesize
\begin{tabular}{lrrr}
\hline
\textbf{Approach} & \textbf{Accuracy} & \textbf{Sentence} & \textbf{On unknowns} \\
\hline
\multicolumn{4}{c}{\textbf{Newswire}} \\
\hline
Brill			& 93.9\%	& 28.4\%	& -	\\
Stanford Tagger		& {\bf 97.3\%}	& {\bf 56.8\%}	& {\bf 90.5\%}	\\
\hline
\multicolumn{4}{c}{\textbf{Microblog}} \\
\hline
Brill			& 70.5\%	& 2.54\%	& 13.1\%	\\
Stanford Tagger		& 73.6\%	& 4.24\%	& 26.6\%	\\
\hline
Brill (twitter)		& 78.6\%	& 8.47\%	& 28.5\%	\\
Derczynski/Stanford (twitter) 	& 88.7\%	& 20.3\%	& 72.1\%	\\
Ritter (twitter)	& 88.3\%	& -		& -	\\
Owoputi (twitter)	& 90.4\%	& 22.0\%     & - \\
\hline
\end{tabular}
\caption{POS tagging performance on two corpora: extracts from the Wall Street journal (newswire), and a 25\% part of Ritter's corpus. The latter is tagged with and without in-genre training data. Accuracy is measured at both token and sentence level.}
\label{tab:pos}

\end{table}

To measure the impact that domain adaptation of POS tagging has on NER in tweets, we attempted NER using part of speech tags generated by a Twitter-specific and a generic tagger.
This was done using GATE ANNIE.
The results are shown in Table~\ref{tab:pos-ner-impact}.

\begin{table}
\centering
\footnotesize
\begin{tabular}{lcccccc}
\hline
 & \multicolumn{2}{c}{\textbf{Person}} & \multicolumn{2}{c}{\textbf{Location}} & \multicolumn{2}{c}{\textbf{Organisation}} \\
 & Prec. & Recall & Prec. & Recall & Prec. & Recall \\
\hline
With Twitter PoS tagging	&28	&21	&71	&25	&14	&7		\\
With newswire PoS tagging	&26 &21	&71	&25	&13	&6		\\
\hline
\end{tabular}
\caption{Impact of domain-specific PoS tagging on NER.}
\label{tab:pos-ner-impact}
\end{table}

A small performance benefit is realised, but recall remains low.
It is worth noting that ANNIE is largely gazetteer-based, and so less sensitive to some types of noise, including wrong POS tags.
Balancing this advantage is the limited recall of gazetteers when faced with previously unseen items, visible here.
However, statistical machine learning-based tools do not reach good levels of recall on tweets either.
We did not perform an analysis using the Stanford NER tool and different POS tag sources because this NER tool includes very similar features to those used for POS tagging, and we have already demonstrated the significant impact that in-domain training data has on it (Table~\ref{tab:ner-ritter}).
For further comment on this issue, see the MSM2013 challenge~\cite{rowe2013msm}, where it was noted that many of the top-performing systems included custom POS taggers.

Despite the significant error reduction, current microblog-adapted POS tagging performance is still not as high as on longer, cleaner content, which impacts NER and NEL performance negatively.

\subsection{Normalisation}
\label{sec:normalisation}

Normalisation is a commonly proposed  solution in cases where it is necessary to overcome or reduce linguistic noise~\cite{sproat2001normalization}.
The task is generally approached in two stages: first, the identification of orthographic errors in an input discourse, and second, the correction of these errors.
Example~\ref{ex:norm} shows an original microblog message, including a variety of errors, and the post-normalisation text. 

\begin{example}
\textbf{Source text:} @DORSEY33 lol aw . i thought u was talkin bout another time . nd i dnt see u either !

\textbf{Normalised text:} @DORSEY33 lol \underline{aww} . \underline{I} thought \underline{you} was \underline{talking} \underline{about} another time . \underline{And} \underline{I} \underline{didn't} see \underline{you} either !
\label{ex:norm}
\end{example}

As can be seen, not all the errors can be corrected (\emph{was} ought to be \emph{were}, for example) and some genre-specific slang remains -- though not in a particularly ambiguous sense or grammatically crucial place.
A variety of custom approaches to normalisation has been developed for Twitter. 
In this section, we present state-of-the-art normalisers; following this, we evaluate the effectiveness of two approaches for improving named entity extraction and linking.

\subsubsection{Normalisation Approaches}

Normalisation approaches are typically based on a correction list, edit-distance based, cluster-based, or a mixture of these, with hybrid approaches common.
Gazetteer-based approaches can be used to repair errors on all kinds of words, not just named entities; for example, Han et al.~\cite{han2012automatically} construct a general-purpose spelling correction dictionary for microblogs.
This achieves state-of-the-art performance on both the detection of mis-spelled words and also applying the right correction.
It is also possible to use a heuristic to suggest correct spellings, instead of a fixed list of variations.

In the case of microblogs, it has been found that slang accounts for three times as many mis-taggings as  orthographic errors~\cite{derczynski2013microblog}.
To this end, as well as investigating a heuristic normaliser, we investigate the performance impact that a custom gazetteer-based general-purpose pre-recognition dictionary has on NER.

Our gazetteer approach is constructed from the Ritter dataset's training split.
Creation of the gazetteer is labour-intensive, and the approach is not flexible when taken across domain, across languages or presented with new kinds of slang or nicknames.
Our heuristic normalisation approach is taken from Han and Baldwin~\cite{han2011lexical}. 
It uses a combination of edit distance, phonetic distance (double metaphone), and a small fixed list of slang words.

Normalisation is not a perfect process.
We noted three major types of mistake:

\textbf{Sentiment change}: 
Some words can be corrected to another word with differing sentiment, that is orthographically close. 
Tolerating mistakes within a Levenshtein edit distance of 2 -- a common limit in this task -- allows \emph{in-}, \emph{im-}, \emph{un-} and \emph{an-} prefixes to be stripped, thus reversing the meaning of a word in some cases. 
For example, \emph{unpossible} $\rightarrow$ \emph{possible} and  \emph{untalented} $\rightarrow$ \emph{talented}.

\textbf{IV errors}:  
The core vocabulary for discriminating between IV/OOV can be too small, or have too low a weight over other factors.
Corrections are made to correctly-spelled words as a result, e.g. \emph{roleplay} $\rightarrow$ \emph{replay} and  \emph{armwarmers} $\rightarrow$ \emph{armorers}.

\textbf{Proper name variations}: 
Corrections to the spelling of variations of family names were sometimes made incorrectly.
For example, there are many variations of spellings of English surnames; \emph{Walmesley} may be alternatively represented as \emph{Walmsley} or \emph{Walmesly}.
This can create critical changes in meaning.
From the data, \emph{She has Huntington's} was mis-normalised to \emph{She has Huntingdon's}, reducing the chances of this already minimal-context mention of a disease being correctly recognised and linked, especially given the reliance of many entity linking techniques upon textual context.

\begin{table}
\centering
\footnotesize
\begin{tabular}{lccc}
\hline
\textbf{Entity type} & \textbf{No norm} & \textbf{Gazetteer norm} & \textbf{Heuristic norm} \\
\hline
Organisation	& 28.6\%	& 28.6\%	& \textbf{29.6\%} 	\\
Location	& 60.5\%	& 60.5\%	& \textbf{62.6\%}	\\
Person		& 63.2\% 	& \textbf{63.6\%}	& 62.8\%	\\
\hline
Overall		& 49.1\%	& 47.6\% 	& \textbf{49.3\%}	\\
\hline
\end{tabular}
\caption{Impact of gazetteer and machine-learning based normalisation techniques on NER F1 measure when extracting named entities, using Stanford NER with a Twitter-trained model.}
\label{tab:norm-findings}
\end{table}

\subsubsection{Impact of Microblog Normalisation on NER Performance}

We measured the performance impact of basic and strong normalisation over a machine-learning based NER system (Stanford CRF NER). 
Results are given in Table~\ref{tab:norm-findings}.

The Stanford NER system relies on context, and so might be more affected by normalisation, which is reflected in the reported results. 
In fact, this system only realises NER performance benefits from strong normalisation. 
While per-category F1 scores are roughly the same (or even slightly improved) with basic normalisation, entities of other types are missed more frequently than without normalisation, leading to an overall decrease in performance.
Use of full normalisation did not cause any more entities to be missed than the no-normalisation baseline, but did slightly reduce the false positive rate.

This motivates our conclusion that while certainly helpful, normalisation is not sufficient to be the sole method of combatting microblog noise.

\section{Conclusion}\label{sec:conclusion}

This paper demonstrated that state-of-the-art NER and NEL approaches do not perform robustly on ill-formed, terse, and linguistically ``compressed" microblog texts. Some Twitter-specific methods reach F1 measures of over 80\%, but are still far from the state-of-the-art results achieved on newswire. 

Next the paper examined the major causes of this poor performance. Poor capitalisation, in particular, causes very significant drops in NER (and thus NEL) recall, not just on microblog content, but also on lowercased newspaper content. Typographic errors and the ubiquitous occurrence of out-of-vocabulary words in microblogs are also detrimental to precision and recall, but the effect is much less prononunced than that of poor capitalisation. We also demonstrated that shortenings and slang are more pronounced in tweets, but again, their effect is not as significant. Moreover, all machine learning NER and NEL methods suffer from the lack of annotated training microblog content, while entity linking methods also suffer from the lack of sufficient context. 

Language identification, microblog-trained POS tagging, and normalisation were investigated, as methods for reducing the effect of microblog noisiness on NER and NEL performance. While this leads to some improvements, precision and recall still remain low. In ongoing work, we are now experimenting with training a microblog-specific recaser, in order to address the capitalisation problem. 

This experiment notwithstanding, we argue that NER and NEL performance will still suffer, due to the very different kinds of entities that are typical in microblogs, as compared against news texts from the training data. Therefore, our conclusion is that microblog-specific NER and NEL algorithms need to be developed, if we are to reach the same level of precision and recall as we now have on news articles. 

Moreover, specific to entity linking, the surrounding textual context in microblogs is often not sufficient for disambiguation, even for human annotators. 
This makes entity linking in microblogs a particularly challenging, open research problem. 
One approach for combating this problem involves leveraging structured knowledge bases, such as Wikipedia, to improve the semantic annotation of microposts with the hope that injecting more prior knowledge will counterbalance the lack of context. Secondly, additional context could come from other messages belonging to the microblog author. Links posted by the user, and the content of their user profile, also provide additional linguistic context. Thirdly, methods may also bring the author's  friends content as additional context, possibly even linking to other sites, using a multi-layered social network model~\cite{magnani2011ml}. 

Last, but not least, creating more human-annotated training corpora of microblog content will allow better algorithm adaptation and parameter tuning to the specifics of this genre. For example, there are currently fewer than 10,000 tweets annotated with named entity types, which is far from sufficient. One way to address this data challenge is through crowdsourcing such corpora in their entirety~\cite{SABOU14.497.L14-1412}. A cheaper approach could be to integrate crowdsourcing within the NER and NEL algorithms themselves~\cite{Demartini2012-ZenCrowd}. In this way, NE mentions that can be linked automatically and with high confidence to instances in the Linked Open Data cloud, will not need to be shown to the human annotators. 
The latter will only be consulted on hard-to-solve cases, which will improve result quality while also lowering the annotation costs.

The creation of such new corpora will also aid comparative evaluations, since, as demonstrated in this paper, current corpora suffer from a number of shortcomings. Namely, issues arise from incompatible annotation types used in the various corpora, i.e. some cover only the four basic types (PER, LOC, ORG, MISC), whereas others make much finer-grained distinctions. Another problem is that tweet-level entity linking requires corpora that focus purely on entities, without mixing topics as well. One step in that direction is the corpus from the TAC knowledge base population task\footnote{\texttt{http://www.nist.gov/tac/2013/KBP}} (TAC-KBP), which has a NEL task using Wikipedia infoboxes as disambiguation targets. Datasets cover English, Spanish, and Chinese, however, none of these will be microblogs. Only longer textual genres will be covered (newswire, web content, discussion fora), annotated for persons, organisations, and geo-political entities.  At the very least, this would be a useful resource for initial training on web and forum content, although microblog-specific data will still be critical.

\section*{Acknowledgments}

The authors thank Roland Roller and Sean McCorry of the University of Sheffield, and the CrowdFlower workers, for their help in annotating the entity-linked dataset; and the reviewers for their constructive remarks.

This work was partially supported by the UK EPSRC grants Nos. EP/I004327/1 and EP/K017896/1 uComp,\footnote{\texttt{http://www.ucomp.eu}} and by the European Union under grant agreements No. 611233 {\sc Pheme},\footnote{\texttt{http://www.pheme.eu}} No. 287863 TrendMiner,\footnote{\texttt{http://www.trendminer-project.eu/}} No. 287911 LinkedTV,\footnote{\texttt{http://www.linkedtv.eu}} and No. 316404 NewsReader.\footnote{\texttt{http://www.newsreader-project.eu}}

\bibliographystyle{elsarticle-num}
\bibliography{ner_extended}

\section*{Vitae}

\textbf{Leon Derczynski} is a post-doctoral Research Associate, who completed a PhD in Temporal Information Extraction at the University in Sheffield under an enhanced EPSRC doctoral training grant. His main interests are in data-intensive approaches to computational linguistics, specialising in information extraction, spatiotemporal semantics, and handling noisy linguistic data, especially social media. He has been working in commercial and academic research on NLP and IR since 2003 with focus on temporal relations, temporal and spatial information extraction,  semantic annotation, usability and social media. 

\textbf{Diana Maynard} is a Research Fellow at the University of Sheffield, UK. She has a PhD in Automatic Term Recognition from Manchester Metropolitan University, and has been involved in research in NLP since 1994. Her main interests are in Information Extraction, opinion mining, social media and Semantic Web technology. Since 2000 she has led the development of the University of Sheffield's opensource multilingual Information Extraction tools, and has led research teams on a number of UK and EU projects. She is chair of the annual GATE training courses, teaches modules on Advanced Information Extraction, opinion mining and social media analysis, and leads the GATE consultancy on Information Extraction and opinion mining. 

\textbf{Giuseppe Rizzo} is a post-doctoral Researcher at the Computer Science Department of the Universit\`{a} di Torino and he is an associate Research Fellow at the Multimedia Communications Department of EURECOM. His main research interests include Information Extraction, Social Media, and Semantic Web. He is a devoted Web evangelist, fascinated by smart web applications and knowledge extraction techniques. Currently Giuseppe is involved in the following projects: ATMS (Advanced Territory Monitoring System) and LinkedTV; in the recent past has collaborated in the OpenSem and EventMedia. 

\textbf{Marieke van Erp} is a postdoctoral researcher on the NewsReader project at VU University. Within the NewsReader project, she is trying to reconstruct news stories from daily news streams. By extracting the who, what, where and when of each article and merging it with previously extracted knowledge, professional decision makers will be provided with a condensed but complete picture. From October 2009 until March 2013, she worked on the Agora project, where she worked on facilitating and assessing the impact of digitally mediated public history in collaboration with Rijksmuseum and the Dutch Institute for Sound and Vision. Dr. van Erp is particularly interested in the interaction between Natural Language Processing and Semantic Web and reproducibility of research results.

\textbf{Genevieve Gorrell} is a researcher with the Natural Language Processing research group at the University of Sheffield, where she works on the GATE open source text analysis software and the TrendMiner FP7 project. Current research interests include named entity recognition and linking, machine learning for named entity disambiguation and biomedical text mining. She received her PhD in 2008 from Link\"{o}ping University for work focused on the development of an efficient incremental algorithm for dataset dimensionality reduction.

\textbf{Rapha\"el Troncy} is now the primary investigator of a number of National (ACAV, DataLift, WAVE) and European (FP7 Petamedia, FP7 LinkedTV, FP7 MediaMixer, CIP PSP Apps4Europe, AAL ALIAS) projects where semantic web and multimedia analysis technologies are used together. His research interest include Semantic Web and Multimedia Technologies, Knowledge Representation, Ontology Modeling and Alignment and Web Science. Raphaël Troncy is an expert in audio visual metadata and in combining existing metadata standards (such as MPEG-7) with current Semantic Web technologies. He works also closely with the IPTC standardization body and the European Broadcasting Union on the relationship between the News Architecture and the Semantic Web technologies.

\textbf{Johann Petrak} is a research fellow at the University of Sheffield, UK where he works with the Natural Language Processing research group.
His research interests include named entity disambiguation and linking, ontology-based information extraction and semantic search. 

\textbf{Kalina Bontcheva} is a senior research scientist and the holder of an EPSRC career acceleration fellowship, working on text summarisation of social media. Kalina received her PhD on the topic of adaptive hypertext generation from the University of Sheffield in 2001. Her main interests are information extraction, opinion mining, natural language generation, text summarisation, and software infrastructures for NLP. She has been a leading developer of GATE since 1999. Kalina Bontcheva coordinated the EC-funded TAO STREP project on transitioning applications to ontologies, as well as leading the Sheffield teams in TrendMiner, MUSING, SEKT, and MI-AKT projects. She was an area co-chair for Information Extraction at ACL’2010 and demos co-chair at COLING’2008. Kalina is also a demo co-chair of ACL’2014. She also coorganises  and lectures at the week-long, annual GATE NLP summer school in Sheffield, which attracts over 50 participants each year.

\end{document}